\documentclass[letterpaper, 10 pt, conference]{ieeeconf}
\IEEEoverridecommandlockouts 
\usepackage{graphicx}
\usepackage[normalem]{ulem}
\usepackage[flushleft]{threeparttable}

\usepackage{amsmath,amsfonts,amssymb}
\usepackage{algorithmic}
\usepackage{algorithm}
\usepackage{array}
\usepackage[caption=false,font=footnotesize]{subfig}
\usepackage{textcomp}
\usepackage{stfloats}
\usepackage{url}
\usepackage{verbatim}
\usepackage{siunitx}

\newcommand{\vect}[1]{\ensuremath{\underrightarrow{#1}}}

\usepackage[backend=bibtex,bibstyle=ieee,citestyle=numeric-comp]{biblatex}
\title{\LARGE \bf Gaussian-Sum Filter for Range-based 3D Relative Pose Estimation in the Presence of Ambiguities}

\author{Syed S. Ahmed, Mohammed A. Shalaby,
Charles C. Cossette, Jerome Le Ny, and James R. Forbes%
\thanks{This work was supported by the NSERC Discovery and Alliance Grant programs, and the Canadian Foundation for Innovation (CFI) program.}
\thanks{S. S. Ahmed, M. A. Shalaby, C. C. Cossette, and J. R. Forbes are with the Department of
Mechanical Engineering, McGill University, 817 Sherbrooke St. W., Montreal, QC
H3A 0C3, Canada. Jerome Le Ny is with the Department of Electrical Engineering, Polytechnique Montreal, Montreal, QC H3T 1J4, Canada. \{{\tt\small syed.shabbir.ahmed@mail.mcgill.ca\}}.}%
}

\newcommand{\ignore}[1]{}  

\newcommand{\mc}[1]{\ensuremath{\mathcal{#1}}}

\newcommand{\mbc}[1]{\ensuremath{\boldsymbol{\mathcal{#1}}}}
\DeclareMathAlphabet{\mbf}{OT1}{ptm}{b}{n}

\newcommand{\mbfdot}[1]{\ensuremath{\dot{\mbf{#1}}}}

\newcommand{\mbfbar}[1]{\ensuremath{\bar{\mbf{#1}}}}
\newcommand{\mbfhat}[1]{\ensuremath{\hat{\mbf{#1}}}}
\newcommand{\mbfcheck}[1]{\ensuremath{\check{\mbf{#1}}}}

\newcommand{\mbfdel}[1]{\ensuremath{\delta{\mbf{#1}}}}
\newcommand{\mbftilde}[1]{\ensuremath{\tilde{\mbf{#1}}}}

\newcommand{\norm}[1]{\ensuremath{\left\Vert#1\right\Vert}}

\newcommand{\trans}{{\ensuremath{\mathsf{T}}}}

\newcommand{\diag}{{\ensuremath{\mathrm{diag}}}}

\newcommand{\utimes}{{\raisebox{-0.6ex}{\kern-1.0ex\raisebox{0.6ex}{\small$\mathsf{v}$}}}}

\newcommand{\rnums}{\mathbb{R}}

\newcommand{\bma}[1]{\left[\begin{array}{#1}}  
\newcommand{\ema}{\end{array}\right]}
\newcommand{\bbm}{\begin{bmatrix}}  
\newcommand{\ebm}{\end{bmatrix}}
\newcommand{\beq}{\begin{equation}}  
\newcommand{\eeq}{\end{equation}}
\newcommand{\bdis}{\begin{displaymath}}  
\newcommand{\edis}{\end{displaymath}}
\newcommand{\beqarray}{\begin{eqnarray}}
\newcommand{\eeqarray}{\end{eqnarray}}
\newcommand{\beqarraynn}{\begin{eqnarray*}}
\newcommand{\eeqarraynn}{\end{eqnarray*}}

\newcommand{\SO}[1]{\ifmmode SO(#1)\else $SO(#1)$\fi}
\newcommand{\SE}[1]{\ifmmode SE(#1)\else $SE(#1)$\fi}

\newcommand{\so}[1]{\ifmmode \mathfrak{so}(#1)\else $\mathfrak{so}(#1)$\fi}
\newcommand{\se}[1]{\ifmmode \mathfrak{se}(#1)\else $\mathfrak{se}(#1)$\fi}

\DeclareMathOperator{\Adj}{\mbf{Ad}}

\newcommand{\dcmspace}{\hspace{0.1em}}
\NewDocumentCommand{\dcm}{ O{} }{\mbf{C}_{#1} {\ifthenelse{\equal{#1}{}}{}{\dcmspace}}}
\NewDocumentCommand{\dcmbar}{ O{} }{\mbfbar{C}_{#1} {\ifthenelse{\equal{#1}{}}{}{\dcmspace}}}
\NewDocumentCommand{\dcmcheck}{ O{} }{\mbfcheck{C}_{#1} {\ifthenelse{\equal{#1}{}}{}{\dcmspace}}}
\NewDocumentCommand{\dcmhat}{ O{} }{\mbfhat{C}_{#1} {\ifthenelse{\equal{#1}{}}{}{\dcmspace}}}
\NewDocumentCommand{\dcmdot}{ O{} }{\mbfdot{C}_{#1} {\ifthenelse{\equal{#1}{}}{}{\dcmspace}}}

\newcommand{\posespace}{\,}
\NewDocumentCommand{\pose}{ O{} O{} }{\mbf{T}_{#2}^{\ifthenelse{\equal{#1}{}}{}{\posespace} #1 \ifthenelse{\equal{#1}{}}{}{\posespace}}}
\NewDocumentCommand{\posebar}{ O{} O{}}{\mbfbar{T}_{#2}^{\ifthenelse{\equal{#1}{}}{}{\posespace} #1 \ifthenelse{\equal{#1}{}}{}{\posespace}}}
\NewDocumentCommand{\posecheck}{ O{} O{} }{\mbfcheck{T}_{#2}^{\ifthenelse{\equal{#1}{}}{}{\posespace} #1 \ifthenelse{\equal{#1}{}}{}{\posespace}}}
\NewDocumentCommand{\posehat}{ O{} O{} }{\mbfhat{T}_{#2}^{\ifthenelse{\equal{#1}{}}{}{\posespace} #1 \ifthenelse{\equal{#1}{}}{}{\posespace}}}
\NewDocumentCommand{\posedot}{ O{} O{} }{\mbfdot{T}_{#2}^{\ifthenelse{\equal{#1}{}}{}{\posespace} #1 \ifthenelse{\equal{#1}{}}{}{\posespace}}}

\begin{document}

%
%
%
%
%
%
%
\def \myJournal {IEEE Conference on Control Technology and Applications}
\def \myDoi {10.1109/CCTA60707.2024.10666532}
\def \myPaperSiteName {IEEE Xplore}
\def \myPaperSiteLink {https://ieeexplore.ieee.org/document/10666532}
\def \myYear {2024}
\def \myPaperCitation{S. S. Ahmed, M. Shalaby, C. C. Cossette, J. Le Ny and J. R. Forbes, ``Gaussian-Sum Filter for Range-based 3D Relative Pose Estimation in
the Presence of Ambiguities,'' in \textit{IEEE Conference on Control Technology and Applications}, August 2024.}


\begin{figure*}[t]

\thispagestyle{empty}
\begin{center}
\begin{minipage}{6in}
\centering
This paper has been accepted for publication in \emph{\myJournal}. 
\vspace{1em}

This is the author's version of an article that has, or will be, published in this journal or conference. Changes were, or will be, made to this version by the publisher prior to publication.
\vspace{2em}

\begin{tabular}{rl}
DOI: & \myDoi\\
\myPaperSiteName: & \texttt{\myPaperSiteLink}
\end{tabular}

\vspace{2em}
Please cite this paper as:

\myPaperCitation

\vspace{15cm}
\copyright \myYear \hspace{4pt}IEEE. Personal use of this material is permitted. Permission from IEEE must be obtained for all other uses, in any current or future media, including reprinting/republishing this material for advertising or promotional purposes, creating new collective works, for resale or redistribution to servers or lists, or reuse of any copyrighted component of this work in other works.

\end{minipage}
\end{center}
\end{figure*}
\newpage
\clearpage
\pagenumbering{arabic} 

\maketitle
\thispagestyle{empty}
\pagestyle{empty}

\begin{abstract}
   Multi-robot systems must have the ability to accurately estimate relative states between robots in order to perform collaborative tasks, possibly with no external aiding. Three-dimensional relative pose estimation using range measurements oftentimes suffers from a finite number of non-unique solutions, or \emph{ambiguities}. This paper: 1) identifies and accurately estimates all possible ambiguities in 2D; 2) treats them as components of a Gaussian mixture model; and 3) presents a computationally-efficient estimator, in the form of a Gaussian-sum filter (GSF), to realize range-based relative pose estimation in an infrastructure-free, 3D, setup. This estimator is evaluated in simulation and experiment and is shown to avoid divergence to local minima induced by the ambiguous poses. Furthermore, the proposed GSF outperforms an extended Kalman filter, demonstrates similar performance to the computationally-demanding particle filter, and is shown to be consistent.
\end{abstract}

\section{Introduction}
The \emph{relative pose}, denoting the relative position and attitude between robots, needs to be accurately estimated to realize autonomous multi-robot tasks. Relative pose information between the robots allows for tasks such as collaborative mapping, collision avoidance, and formation control. Sensors such as cameras with object-detection ability \cite{Li2022Localization} or LiDARs \cite{Cao2020virslam} can satisfy the relative pose estimation requirement, but they are computationally expensive.

Ultra-wideband transceivers, or UWB \emph{tags}, are small, inexpensive range sensors providing approximately $10\,\si{cm}$ accurate range (distance) measurements between a pair of tags for relative position estimation. Inter-robot range data from these tags, fused with other sensor data, enable infrastructure-free robot localization. For instance, relative robot poses can be estimated by fusing velocity estimates from visual-inertial odometry (VIO) with range measurements from a single tag on each robot \cite{Xu2020VU}. This method requires the robots to be in persistent relative motion \cite{Yanjun2020SingleUwb,Hepp2016PersonTO,Samet2018,Charles2021RelPos, Xu2022} or in periodic line-of-sight of the cameras for visual recognition, which is not possible in a forest environment, for example.

Multiple tags can be installed on each robot for relative position estimation \cite{Guler2018LO,Nguyen2018RTL, Cao2018DynamicRL, Fishberg2022}. In fact, placing two UWB tags per robot ensures ``local observability'' \cite{Shalaby2021RP}, thus overcoming the persistent relative motion requirement for position estimation using an extended Kalman filter (EKF) \cite{Shalaby2021RP,Charles2022OptimalMF}. This setup combined with an interoceptive IMU or velocity readings allow for infrastructure-free relative pose estimation in 3D. However, even with two tags per robot, the range measurements yield multiple solutions for relative robot poses, referred to as ``discrete''  \emph{ambiguities}, which are not addressed in \cite{Shalaby2021RP}. 

These ambiguities form a multi-modal distribution of relative poses that the estimator must account for. Adding more UWB tags per robot reduces the number of ambiguities at the cost of the tags not communicating at their highest data rate. In fact, even with three strategically-positioned tags, only relative robot positions can be disambiguated, while relative attitude still remains ambiguous. Therefore, for a range-based approach, designing estimators that can handle these ambiguities is of great importance.

In the face of ambiguities, Gaussian-based filters, such as an EKF, can perform poorly since they assume that the distribution is unimodal \cite{Kamthe2014}. A particle filter (PF) can handle a  multi-modal distribution \cite{Maskell2013OptimisedPF,Bilgin2015MultiMP,Nguyen2008}, but it is computationally expensive due to the need for many particles to describe the multi-modality \cite{Terejanu2008}. Range-based localization of the ambiguous position of one robot with three static anchors in 2D has been addressed using a Gaussian-sum filter (GSF) in \cite{Kim2020}. Additionally, signal map measurements often exhibit multi-modality while tracking multiple targets \cite{Raitoharju2020GaussianMM}, and a Gaussian mixture model (GMM) helps isolate the ``true'' measurement for a particular target. In this paper, the ideas presented in \cite{Kim2020,Raitoharju2020GaussianMM} are extended to design a localization solution involving a Gaussian-sum filter (GSF) where the ``true'' relative pose between multiple robots is identified among the ambiguous poses in 3D. Unlike \cite{Kim2020}, this solution only uses two UWB tags per robot, no static anchors, and provides a complete 3D pose estimation solution that includes both position and attitude.

As such, the key contributions of this paper are as follows.
\begin{itemize}
    \item Identification of all the possible ambiguous relative poses between $N$ robots using a geometric approach is presented. The geometric estimates are fed into a least-squares estimator to form a GMM of ambiguous relative poses in 3D. These estimates are used to initialize a GSF to identify the ``true'' relative pose. Since this GSF is only initialized at the ambiguous poses, it contains the minimum number of Gaussian components required to model the multi-modal state.
    \item To the best of the Author's knowledge, this is the first work where a GSF is used for anchor-free, range-based 3D relative pose estimation between robots in the presence of ambiguities. 
    \item In simulations and experiments, the proposed estimator involving the GSF is shown to have a similar performance to the PF, while, as expected, being orders of magnitude faster.
\end{itemize}

The remainder of this paper is organized as follows. The notation and preliminaries are defined in Section \ref{sec:problem_setup}. The problem formulation is in Section \ref{sec:estimation} and the GSF is discussed in Section \ref{sec:gsf}. The ambiguous pose estimation procedure for initializing the GSF is presented in Section \ref{sec:initialization}. The estimator is validated in simulation and experiment in Sections \ref{sec:simulation} and \ref{sec:experimental}, respectively.

\section{Notation and Preliminaries}\label{sec:problem_setup}    
\begin{figure}[b]
    \vspace{-0.11cm}
    \centering
    \subfloat[{Problem setup}\label{fig_1}]{{\includegraphics[width=1.25in, trim={0cm, 0cm, 0cm, 0.5cm}, clip]{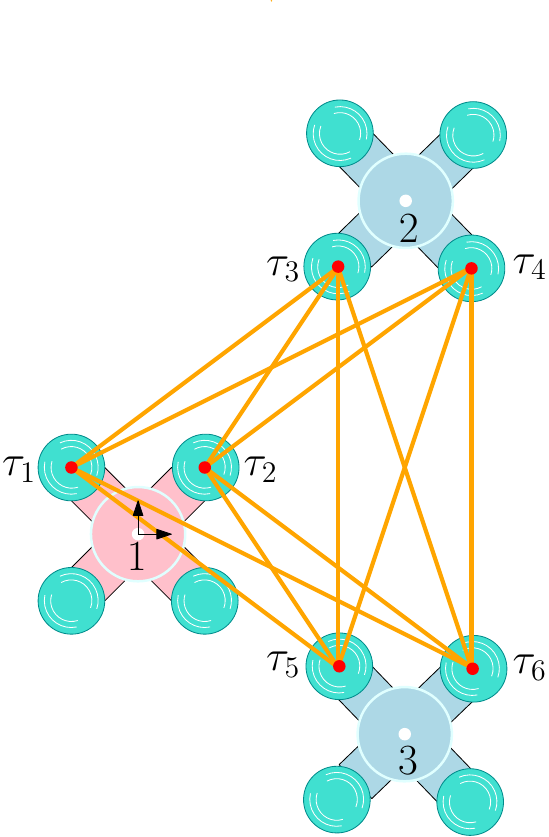}}}%
    \hspace{0.6cm}
    \subfloat[{Ambiguous relative poses}\label{fig_2}]{{\includegraphics[width=1.65in, trim={0cm 0cm 0cm 0.5cm}, clip]{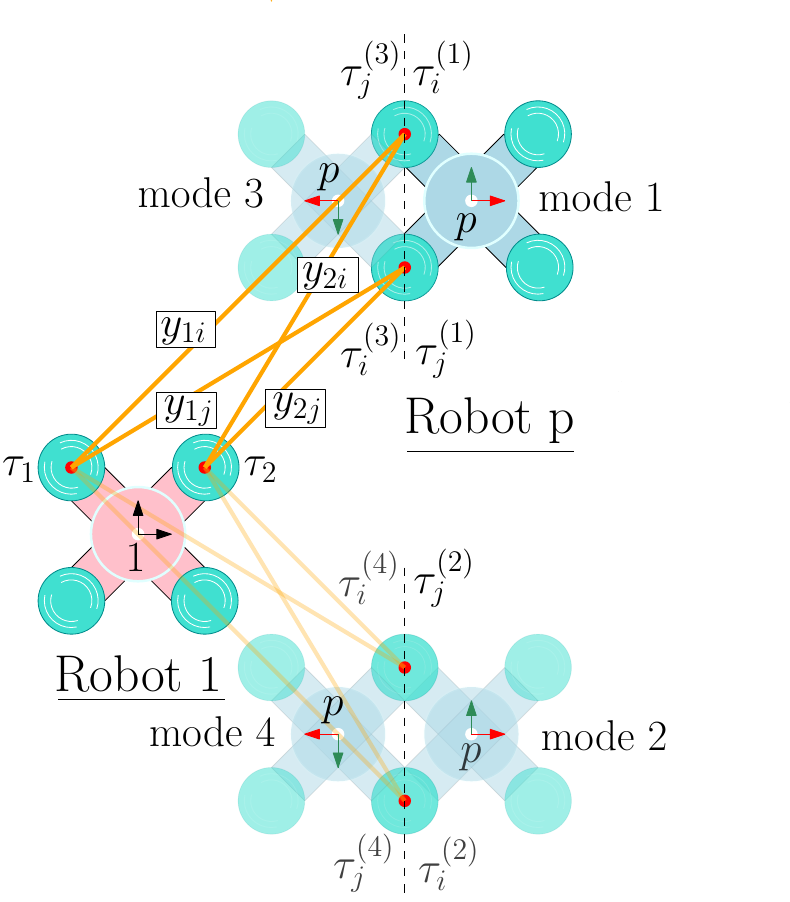} }}%
    \caption{\hspace{-0.15cm}(a) Problem setup for a two-tag multi-robot system, where each robot is equipped with Tags $\tau_i$ and $\tau_j$. Without loss of generality, the pink robot, defined as Robot~$1$, is considered to be the reference robot.     
  \newline(b) Visualization of all the possible ambiguous relative poses between robots $1$ and $p$. The relative pose in mode $1$ is the ``true'' pose and modes $2,\,3,$ and $4$ are ambiguities. The range measurements are $y_{1i},\,y_{1j},\,y_{2i},$ and $y_{2j}$.}%
    \label{fig_0}
\end{figure}

Consider $N$ robots or rigid bodies, where each robot is equipped with two ranging tags, resulting in a total of $2N$ tags collectively, as shown in Fig.~\ref{fig_1}. The tags are located at physical points $\tau_1, \ldots, \tau_{2N}$ on the robots. A measurement graph $\mc{G} = (\mc{V}, \mc{E})$ denotes the inter-tag range measurements. The nodes $\mc{V} = \{1, \ldots, 2N\}$ are the set of tag IDs and the edges $\mc{E}$ denote the set of inter-tag range measurements. The bolded $\mbf{1}$ and $\mbf{0}$ are appropriately-sized identity and zero matrices, respectively.

An orthonormal reference frame $\mc{F}_p$ is attached to each robot, where $p = 1, \ldots, N$ are the robot IDs. $\mc{F}_g$ is defined as a common global reference frame, and $w$ is a static point. The position of a chosen reference point in Robot $p$ relative to point $w$, resolved in $\mc{F}_p$ is denoted $\mbf{r}^{pw}_p \in \rnums^n$, and the robot's translational velocity with respect to another arbitrary reference frame $\mc{F}_c$ is denoted $\mbf{v}^{pw/c}_p \in \rnums^n$. Vectors resolved in different frames are related by the transformation, $\mbf{r}_p^{pw}~=~\mbf{C}_{pq} \mbf{r}_q^{pw}$, where $\mbf{C}_{pq} \in \SO{n}$, where $\SO{n}$ is the special Orthogonal group in $n$ dimensions. The angular velocity of $\mc{F}_p$ relative to $\mc{F}_q$ resolved in $\mc{F}_c$ is denoted $\mbc{\omega}^{pq}_c$.
The relative pose between Robots $p$ and $q$ is 
\begin{align}
    \mbf{T}_{pq} = \bbm
    \mbf{C}_{pq} & \mbf{r}^{qp}_p \\
    \mbf{0} & 1
    \ebm  \in \SE{n},
\end{align}
where $\SE{n}$ is the special Euclidean group in $n$ dimensions. The exponential map of $\SE{n}$ is denoted $\exp: \se{n} \rightarrow \SE{n}$, where $\se{n}$ is the Lie algebra of $\SE{n}$. The ``wedge'' operator is denoted $(\cdot)^\wedge: \rnums^m \rightarrow \se{n}$, and the ``vee'' operator is $(\cdot)^\vee: \se{n} \rightarrow \rnums^m$. The adjoint matrix is denoted $\mbf{Ad}: SE(n) \rightarrow \rnums^{m\times m}$ and is defined in \cite[Pg.~324]{ Barfoot2017StateEF}. The $\mbf{a}^\odot$ operation is given in \cite[Pg.~310]{Barfoot2017StateEF}.

\section{Problem Formulation} \label{sec:estimation}
The poses of all the robots are expressed relative to Robot~$1$, which is the arbitrarily-chosen reference robot. The state of the system is defined as,
\begin{align}\
    \label{eq:state_def}
  \mbf{x} = (\mbf{T}_{12}, \ldots, \mbf{T}_{1N}) \in \SE{n}^{N-1}.
\end{align}
With $\delta\mbc{\xi}_p \in \rnums^{m}$, $\mbfdel{x} = [\delta\mbc{\xi}_2^\trans \cdots \delta\mbc{\xi}_N^\trans]^\trans \in \rnums^{m(N-1)}$, the $\oplus$ operator is defined as
\begin{align}
    \label{eq:oplus}
    \mbf{x} \oplus \delta \mbf{x} = (\mbf{T}_{12}\exp(\delta\mbc{\xi}_2^\wedge), \ldots, \mbf{T}_{1N}\exp(\delta\mbc{\xi}_N^\wedge)).
\end{align}

The objective is to accurately estimate the state $\mbf{x}$. For this, the interoceptive measurements are each robot's angular and translational velocities as resolved in its body frame, denoted as
\begin{align}
    \mbf{u}_{p} = [
    \mbc{\omega}^{pg\trans}_p \;\;
    \mbf{v}^{pw/g\trans}_p]^\trans
    + \mbf{w}_p \in \rnums^m, \quad \mbf{w}_p \sim \mc{N}(\mbf{0},\mbf{Q}_p), \nonumber
\end{align}
where $\mbf{w}_p$ is zero-mean Gaussian noise with covariance $\mbf{Q}_p$. The relative pose between Robots $1$ and $p$ at time-step $k$ is
\begin{align}
    \mbf{T}_{1p_k} &= \mbf{T}_{g1_k}^{-1} \mbf{T}_{gp_k} \nonumber \\
                   &= \Bigl(\exp(-\Delta t\,\mbf{u}_{1_{k-1}}^\wedge) \mbf{T}_{g1_{k-1}}^{-1}\Bigr) \Bigl(\mbf{T}_{gp_{k-1}}  \exp(\Delta t\,\mbf{u}_{p_{k-1}}^\wedge)\Bigr) \nonumber \\
                   \label{eq:process_model}
                   &\triangleq \mbf{f}(\mbf{T}_{1p_{k-1}}, \mbf{u}_{1_{k-1}}, \mbf{u}_{p_{k-1}}),
\end{align}
where $\Delta t = t_k - t_{k-1}$ is the time interval. The relative poses $\mbf{T}_{1p_k},\, p= 2, \ldots, N,$ collectively form the state $\mbf{x}_k$.

Meanwhile, a range measurement between Tags $i$ and $j$ in Robots $p$ and $q$ respectively, is modelled as
\begin{align} \label{eq:meas_model}
    y_{ij} (\mbf{x}_k) &= \norm{ \mbf{D}\mbf{T}_{1p_k}
    \mbftilde{r}^{\tau_i p}_p
    - \mbf{D} \mbf{T}_{1q_k}
    \mbftilde{r}^{\tau_j q}_q
     } + \eta_{ij},
\end{align}
where $\mbf{D} = [\mbf{1}\; \mbf{0}]$,
$\mbftilde{r} = [\mbf{r}^{\trans} \;1]^\trans$, $||\cdot||$ is the Euclidean norm, and $\eta_{ij} \sim \mc{N}(0, \sigma^2_{ij})$. The augmented measurement vector is
\begin{align}
    \label{eq:meas_model2}
    \mbf{y}_k &= \mbf{g}_k(\mbf{x}_k) + \mbc{\eta}_k =[\cdots y_{ij}(\mbf{x}_k) \cdots]^\trans + \mbc{\eta}_k \in \rnums^{|\mc{E}|}, \nonumber \\
    &\hspace*{3.4cm}\forall (i,j) \in \mc{E}, \mbc{\eta}_k \sim \mc{N}(\mbf{0}, \mbf{R}_k), 
\end{align}
where $\mbf{R}_k=\diag(\ldots, \sigma^2_{ij}, \ldots)$.

Estimating the state $\mbf{x}$ of a multi-robot system with two tags per robot is non-trivial. As shown in Fig.~\ref{fig_2}, in this setup, there is a finite set of discrete relative poses or ambiguities that correspond to the same range measurements. These ambiguities will be referred to as \emph{modes} in the paper.
In 2D, the two obvious ambiguities are modes 1 and 2 since the range measurements are equal in both the modes. Given noisy range measurements, when $y_{1i} \approx y_{1j}$ and $y_{2i} \approx y_{2j}$, there is a likelihood of ``flip'' ambiguities occurring, where tags $\tau_i$ and $\tau_j$ swap their positions, yielding modes 3 and 4. These modes present an issue for estimator initialization when robots are static, as there is no motion to disambiguate 
the multiple modes. 

The multi-modal state representing this system can be estimated using a GSF. The GSF is typically initialized by sampling from either a uniform distribution of all possible states or a Gaussian distribution based on the prior knowledge. With limited prior knowledge, both methods can require many Gaussian components in the GSF. In this paper, a Gaussian component is assigned per ambiguous pose in 2D, which are denoted as modes in Fig. \ref{fig_2}, to form a GMM that captures the state's multi-modality effectively. This GMM is used to initialize a GSF that isolates the ``true'' mode when the robots are in motion and avoids divergence to new ambiguities in-flight, which allows accurate and efficient state estimation. This novel initialization method minimizes the number of Gaussian components required in the GSF, thus improving computational efficiency. 




To initialize the GSF using the proposed GMM, given the challenge posed by measurement noise, a two-step solution is undertaken. Firstly, analytical geometric derivations are used to evaluate all ambiguous poses in 2D as a preliminary guess. Secondly, this guess is refined through a nonlinear least-squares algorithm to get a more accurate estimate. These methodologies are discussed in Section \ref{sec:initialization}.

\section{Gaussian-Sum Filter} \label{sec:gsf}
The GSF is a Bayesian filter that approximates the state distribution by a weighted sum of Gaussian probability density functions \cite{Alspach1972GSF}. The GSF consists of $M$ EKFs, each initialized at time-step $k_0$ with an equal weightage at a different initial state, $\mbfcheck{x}^{(i)}_0$ and covariance, $\mbfcheck{P}^{(i)}_0$, $i = 1,\ldots,M$, such that the weights sum to $1$, which is stated as $\sum_{i=1}^M w^{(i)}_{0} = 1$. Each of these EKFs is referred to as a \emph{mode} of the GSF. 

The evaluation of the modes of the GSF is done using an EKF prediction and correction step, which are detailed in \cite{Barfoot2017StateEF}. The process and measurement models are given in \eqref{eq:process_model} and \eqref{eq:meas_model2}, respectively. The process model Jacobian, $\mbf{A}(\mbf{x})$, and the measurement model Jacobian, $\mbf{H}(\mbf{x})$, are given in Section \ref{sec:process_jacobian} and Section \ref{sec:meas_jacobian}, respectively. 

The EKFs are run in parallel and the posterior density at time-step $k$ is represented as a Gaussian sum of $M$ modes, with the $i^\text{th}$ mode weighted using $w^{(i)}_{k}$. The primary feature of the GSF is that when a new measurement $\mbf{y}_k$ is received, the weights are updated by comparing the measurement with the predicted measurement of each mode, which is given by 
\begin{align}
    \mbfcheck{y}_k^{(i)} = \mbf{g}(\mbfcheck{x}_k^{(i)}),
\end{align}
where, $\mbfcheck{x}^{(i)}_k$ is the predicted states of the $i^\text{th}$ mode. If a mode's predicted value $\mbfcheck{y}^{(i)}_k$ closely matches $\mbf{y}_k$, it is more likely to be responsible for the observation and thus receives a higher weight and vice versa. As shown in \cite{Alspach1972GSF}, the weights quantify the probability of a measurement being associated with each mode, and are updated as
\begin{align}
    w_k^{(i)} = \frac{w_{k-1}^{(i)}\mc{N}(\mbf{y}_k; \mbf{g}(\mbfcheck{x}_k^{(i)}), \mbf{S}_k^{(i)})}{\sum_{i=1}^M w_{k-1}^{(i)}\mc{N}(\mbf{y}_k; \mbf{g}(\mbfcheck{x}_k^{(i)}), \mbf{S}_k^{(i)})},
\end{align}
where $\mbf{S}_k^{(i)}$ is the covariance matrix of the predicted measurement of the $i^\text{th}$ mode.
The mean estimate of the GSF is a weighted average of the estimates in all the modes in $\rnums^m$, which updates $\mbfhat{x}_k$ using Gaussian mixture reduction as shown in \cite{CesicMR2017}. The mean in $\rnums^m$ is given by
\vspace{-0.05cm}
\[\begin{array}{l}
    \delta\mbfhat{x}_k = \sum_{i=1}^M w_k^{(i)}\delta\mbfhat{x}_k^{(i)}, 
\end{array}\]
where, as per \cite{CesicMR2017}, $\delta\mbfhat{x}_k^{(i)} = \mbfhat{x}_k^{(i)} \ominus\,\mbfhat{x}^{(\alpha)}_k$, $\alpha = \arg \max_{i} w_k^{(i)}$, and the state is updated as $\mbfhat{x}_k = \mbfhat{x}_{k-1} \oplus \,\delta\mbfhat{x}_k$. The covariance is assumed as the covariance of the max-weighted mode since the objective is to detect the true Gaussian mode, and is given by
\[\begin{array}{l}
\mbfhat{P}_k = \mbfhat{P}_k^{(\alpha)}.
\end{array}\]
For a detailed derivation of the GSF, refer to \cite{Brian1979OF}.

\subsection{Process Model Jacobian} \label{sec:process_jacobian}
Let $\mbf{T}_{1p} = \bar{\mbf{T}}_{1p} \exp(\delta\mbc{\xi}^\wedge)$, where $\bar{\mbf{T}}_{1p} \in SE(n)$, and $\delta\mbc{\xi} \in \rnums^{m}$ is small. Replacing $\mbf{T}_{1p_k}$ and $\mbf{T}_{1p_{k-1}}$ by this approximation into \eqref{eq:process_model} and left-multiplying both sides by $\bar{\mbf{T}}_{1p_k}^{-1}$ yields
\begin{align}
    &\exp(\delta\mbc{\xi}^\wedge_k) = \exp(-\Delta t\,\mbf{u}_{p_{k-1}}^\wedge) 
    \exp(\delta\mbc{\xi}^\wedge_{k-1})\exp(\Delta t\,\mbf{u}_{p_{k-1}}^\wedge). \nonumber
\end{align}
Given, $\exp((\Adj(\mbf{T})\delta\mbc{\xi})^\wedge) \equiv \mbf{T}\exp(\delta\mbc{\xi}^\wedge)\mbf{T}^{-1}$ \cite{Barfoot2017StateEF}, it follows that
$
    \delta\mbc{\xi}_k = \Adj(\exp(-\Delta t\, \mbf{u}_{p_{k-1}}^\wedge))\delta\mbc{\xi}_{k-1}.
$ 
Based on \cite{sola2021micro},
\begin{align}
    \label{eq:process_jacobian2}
\frac{D\mbf{f}(\mbf{T}_{1p_{k-1}}, \mbf{u}_{1_{k-1}}, \mbf{u}_{p_k-1})}{D\mbf{T}_{1p_{k-1}}} &= \Adj(\exp(-\Delta t \,\mbf{u}_{p_{k-1}}^\wedge)).
\end{align}
Thus, $\mbf{A}_{k-1}(\mbf{x})$ a block-diagonal matrix in $\rnums^{(m\times m)(N-1)}$, where the $(p-1)^\text{th}$ block is given by \eqref{eq:process_jacobian2}, for $p = 2, \ldots, N$.

\subsection{Measurement Model Jacobian} \label{sec:meas_jacobian}
As derived in \cite{Charles2022OptimalMF}, the measurement model Jacobian is given by
$\mbf{H}(\mbf{x}) = [\cdots \mbf{H}^{ij}(\mbf{x})^\trans \cdots]^\trans$,
where,
\begin{align}
        \mbf{H}^{ij}(\mbf{x}) &= 
        \Bigl[
        \mbf{0} \cdots \mbf{H}^{ij}_p(\mbf{x}) \cdots \mbf{H}^{ij}_q(\mbf{x}) \cdots \mbf{0}
        \Bigr] &&\in \rnums^{1 \times m(N-1)}, \nonumber \\
        \label{eq:h1}
    \mbf{H}^{ij}_p(\mbf{x}) &= \mbc{\rho}_{ij} \mbf{D}\bar{\mbf{T}}_{1p}\mbftilde{r}^{\tau_i p\,\odot}_p &&\in \rnums^{1 \times m},
    \\
    \label{eq:h2}
    \mbf{H}^{ij}_q(\mbf{x}) &= -\mbc{\rho}_{ij}
    \mbf{D}\bar{\mbf{T}}_{1q}\mbftilde{r}^{\tau_j q\,\odot}_q
&&\in \rnums^{1 \times m}, \\
    \mbc{\rho}_{ij} &= \frac{\mbf{D}\mbf{T}_{1p} \mbftilde{r}^{\tau_i p}_p - \mbf{D} \mbf{T}_{1q}\mbftilde{r}^{\tau_j q}_q}{||\mbf{D}\mbf{T}_{1p} \mbftilde{r}^{\tau_i p}_p - \mbf{D} \mbf{T}_{1q} \mbftilde{r}^{\tau_j q}_q||}.
\end{align}
The $p^\text{th}$ and $q^\text{th}$ block columns of $\mbf{H}^{ij}(\mbf{x})$ are populated by \eqref{eq:h1} and \eqref{eq:h2}, respectively.

\section{GSF Initialization Process}\label{sec:initialization}
\subsection{Pose Evaluation using Geometry} \label{sec:geom_eval}
The estimation of the four possible solutions for relative poses between Robot~$1$ and Robot~$p$ in 2D, as shown in Fig.~\ref{fig_2}, is a challenging problem. They are first computed using a geometric method. These solutions form a combination of all ambiguous relative poses between Robots $1$ to $N$. Since the robots only have two ranging tags each, to ensure a finite number of solutions, the problem is addressed in 2D, assuming zero relative roll, and pitch between the robots. Because the robots can be at different heights, to project the range measurements into a 2D plane, the relative height between the robots is taken from the laser-range finders mounted on the robots, and then a 2D projection is performed on
the range measurements. This method assumes that the robots are static or hovering over a common flat surface, which is a reasonable assumption for indoor environments.


The notational preliminaries are as follows. The Tags $1$ and $2$ are in Robot~$1$ and Tags $i$ and $j$ are in Robot~$p$. The range measurements between Robots~$1$ and $p$ are  $y_{1i}$, $y_{1j}$, $y_{2i}$, and $y_{2j}$. The unit vector between tags $\tau_1$ and $\tau_2$ is,
\begin{align}
\mbf{n}_1 = \frac{1}{d}\mbf{r}^{\tau_2\tau_1}_1, \, d=||\mbf{r}^{\tau_2\tau_1}_1||, \quad 
\text{and} \quad \mbf{n}_{1\perp}=\Big[\begin{smallmatrix}
    0 & -1 \\
    1 &  \;\;\,0
    \end{smallmatrix}\Big]
    \mbf{n}_1,
\end{align}
is its dextral orthonormal counterpart. Additionally, note that, any attitude $\mbf{C}_{pq} \in \SO2$ between the frames $\mc{F}_p$ and $\mc{F}_q$ is a function of the heading $\phi_{qp}$ between the frames, and is denoted as $\mbf{C}_{pq} \triangleq \mbf{C}_{pq}(\phi_{qp})$ \cite{Barfoot2017StateEF}.

\begin{figure}[!h]
    \centering
    \subfloat[{Geometry between tags}\label{fig_geom}]{{\includegraphics[width=1.3in, trim={0cm, 0cm, 0cm, 0cm}, clip]{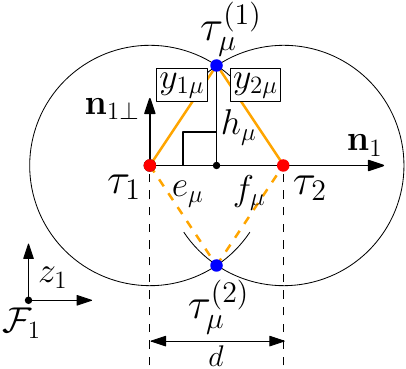}}}%
    \hspace{0.4cm}
    \subfloat[{Geometry between frames}\label{fig_angle}]{{\includegraphics[width=1.8in, trim={0cm, 0cm, 0cm, 0cm}, clip]{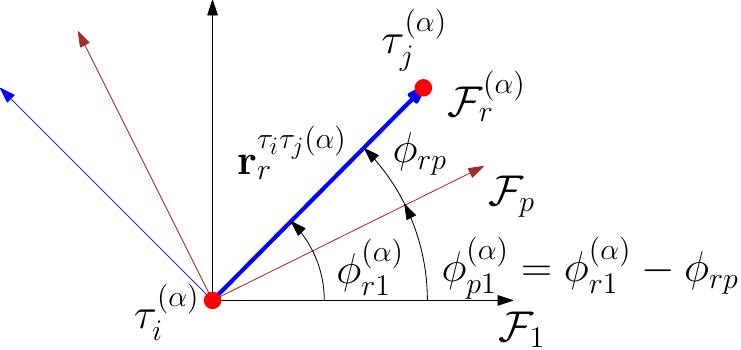} }}%
    \caption{\hspace{-0.3cm}(a) Visualization of the geometric relation between tags $\tau_1$, $\tau_2$ of Robot~$1$ and $\tau_\mu$, $\mu \in \{i,j\}$ of Robot~$p$ resolved in $\mc{F}_1$. The range measurements consist of $y_{1\mu}$ and $y_{2\mu}$, $\mu \in \{i,j\}$. The reference point in Robot~$1$, $1$, and the frame $\mc{F}_1$ are arbitrarily defined.
    (b) Visualization of the relation between frames 
    $\mc{F}_1$, $\mc{F}_p$, and $\mc{F}_r$. Tags $\tau_i$ and $\tau_j$ are mounted on Robot~$p$. In both figures, the superscript $(\cdot)$ represents the mode number. }
    \label{fig_geom_angle1}
    \vspace{-0.01cm}
\end{figure}

In Fig.~\ref{fig_geom}, the two possible position vectors between Tags $\tau_1$ and $\tau_\mu$, $\mu \in \{i,j\}$, and subsequently, the possible position vectors between Tags $\tau_i$ and $\tau_j$ are,
\begin{align}
&e_{\mu} = \frac{1}{2d}(y_{1\mu}^2 - y_{2\mu}^2 + d^2), \;
h_\mu = \surd{(y_{1\mu}^2 - e_\mu^2)}, \; \mu \in \{i,j\}, \nonumber \\
\label{eq:r3}
&\mbf{r}^{\tau_\mu \tau_1(1)}_1
= e_\mu \mbf{n}_1 + h_\mu
\mbf{n}_{1\perp}, \quad \mu \in \{i,j\},\\
&\mbf{r}^{\tau_\mu \tau_1(2)}_1
= e_\mu \mbf{n}_1 - h_\mu
\mbf{n}_{1\perp}, \quad \mu \in \{i,j\}, \\
\label{eq:r4}
&\mbf{r}^{\tau_i\tau_j(\alpha)}_1 = \mbf{r}^{\tau_i\tau_1(\alpha)}_1 - \mbf{r}^{\tau_j\tau_1(\alpha)}_1,\quad \alpha = 1,\,2,
\end{align}
where $\alpha$ is the mode number of the ambiguity. 

A right-handed frame denoted as $\mc{F}_r^{(\alpha)}$ whose $x$-axis is aligned with the physical vector $\vect{r}^{\tau_i\tau_j(\alpha)}$ is shown in Fig.~\ref{fig_angle} in blue. The heading of $\mc{F}_r$ relative to $\mc{F}_p$ and $\mc{F}_1$, and subsequently the attitude, $\mbf{C}_{1p}^{(\alpha)}$, in modes $1$ and $2$ are,
\begin{align}
    \phi_{rp} &= \tan^{-1}(y_p/x_p), \quad \quad &&\text{ s.t.} \;\; \mbf{r}^{\tau_i\tau_j}_p = [x_p \;\; y_p]^\trans, \nonumber \\
    \phi_{r1}^{(\alpha)} &= \tan^{-1}(y_1^{(\alpha)}/x_1^{(\alpha)}), \,\;\,&&\text{ s.t.}
    \;\; \mbf{r}^{\tau_i\tau_j(\alpha)}_1 = [x_1^{(\alpha)} \;\; y_1^{(\alpha)}]^\trans, \nonumber
\end{align}
\begin{align}
    \mbf{C}_{1p}^{(\alpha)} = \mbf{C}_{1r}^{(\alpha)}\mbf{C}_{pr}^\trans, \quad \alpha = 1,\,2. 
\end{align}
Thus, the relative robot positions in modes $1$ and $2$ are,
\begin{align}
    \label{eq:r1}
    \mbf{r}^{p1(\alpha)}_1 = \mbf{C}_{1p}^{(\alpha)} \mbf{r}^{p\tau_i}_p + \mbf{r}^{\tau_i\tau_1(\alpha)}_1 + \mbf{r}^{\tau_11}_1, \quad \alpha = 1,\,2.
\end{align}

The flip ambiguities in modes $3$ and $4$ are reflections of the modes $1$ and $2$ about the axis joining the Tags $\tau_i$ and $\tau_j$ relative to $\mc{F}_1$, given by \cite[Eq. (8)]{Zengin2021},
\begin{align}  
    \mbf{r}^{p1(\alpha+ 2)}_1 = \frac
    {\bbm
    d_\alpha && -2a_\alpha b_\alpha \\
    -2a_\alpha b_\alpha && -d_\alpha
    \ebm \mbf{r}^{p1(\alpha)}_1
    -2c_\alpha
    \bbm
    a_\alpha\\
    b_\alpha
    \ebm}{a_\alpha^2+b_\alpha^2},
    \nonumber
\end{align}
where, $d_\alpha = b_\alpha^2-a_\alpha^2$, $c_\alpha=\diag(-a_\alpha,\,b_\alpha)\,\mbf{r}^{\tau_i1(\alpha)}_1$, and $\mbf{r}^{\tau_j\tau_i(\alpha)}_1 =  [b_\alpha\;a_\alpha]^\trans$, $\alpha = 1,\,2$. As shown in Fig.~\ref{fig_2}, the respective attitudes in these modes have a heading of $\pi$ relative to the attitudes in modes $1$ and $2$, given by,
\begin{align}
\mbf{C}_{1p}^{(\alpha + 2)} = \mbf{C}(\pi)\mbf{C}_{1p}^{(\alpha)}, \quad \alpha = 1,\,2.
\end{align}

By repeating this process, there will be $4$ modes of $N-1$ relative poses between Robots $1$ and $p$, for $p=2,\ldots, N$. Therefore, the total number of combinations of modes are $M = (4)^{N-1}$, collectively denoted as $\mbf{x}^{(i)}_{\text{geom}}$, $i=1,\ldots, M$.

\begin{figure}[!b]
    \vspace{-0.2cm}
    \centering
    \subfloat[Simulated UWB range data\label{fig_sa}]{{\includegraphics[scale = 0.211, trim={9.4cm 0.1cm 11.5cm 2.5cm}, clip]{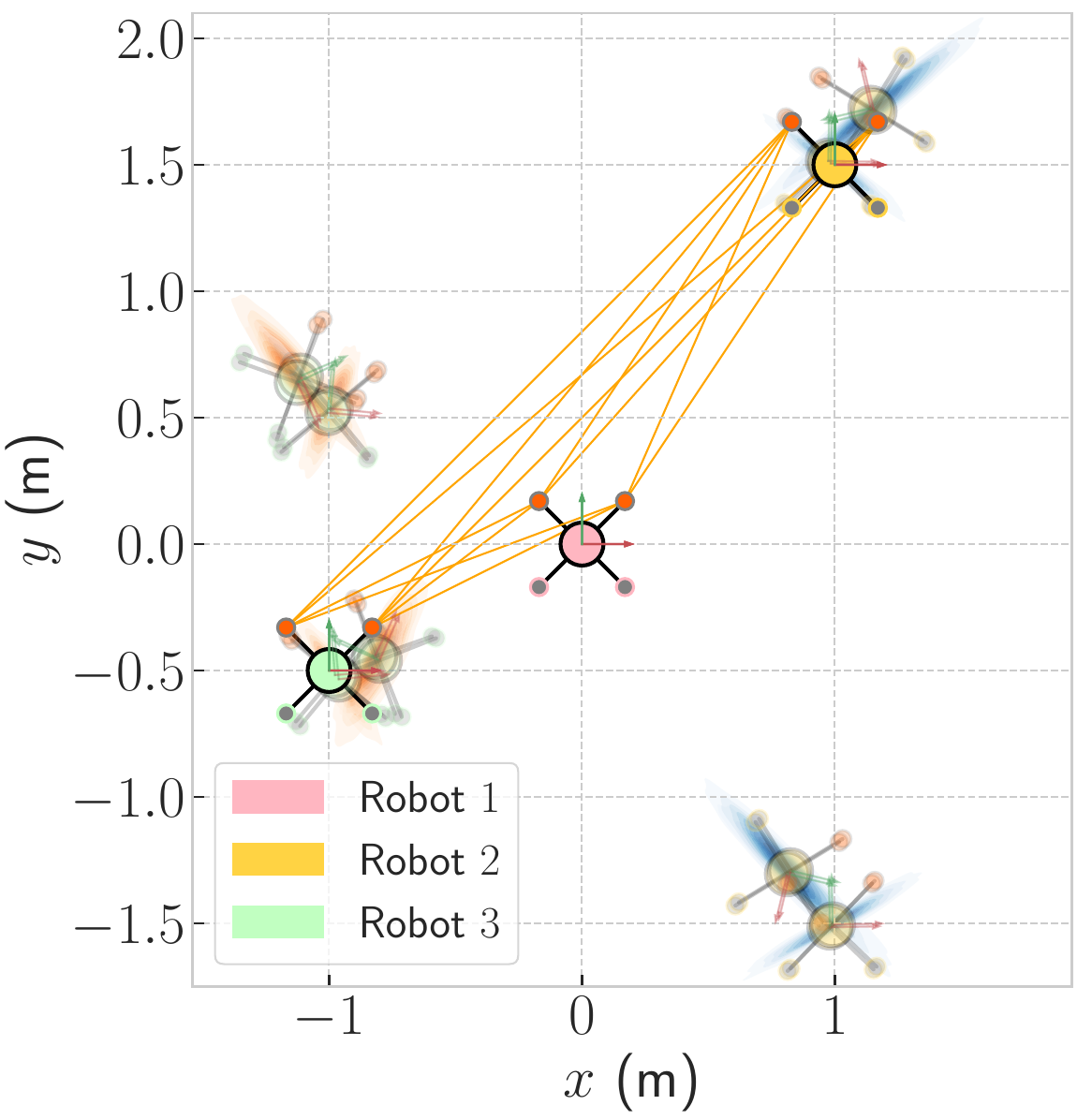}}}%
    \subfloat[Experimental UWB range data]{{\includegraphics[scale = 0.211, trim={9cm 0.12cm 8cm 2.48cm}, clip]{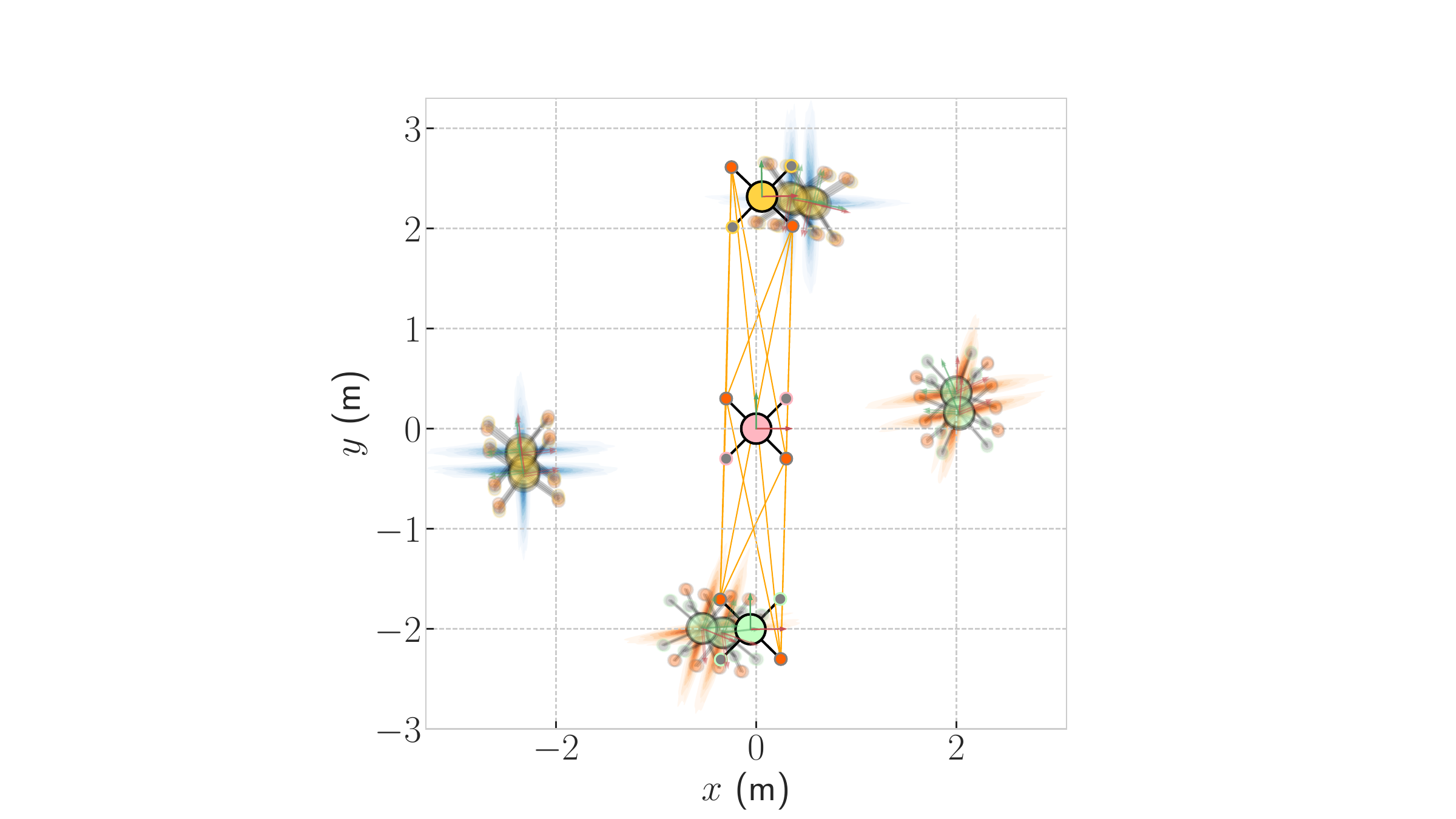} }}%
    \caption{Comparison between the true pose and the ambiguous GI-LS pose estimates in a system of three robots, each having two tags. The opaque drones denote the true poses. The lighter shaded drones with their respective covariance plots are the pose estimates and their corresponding uncertainties.}%
    \label{fig_3}
\end{figure}

\subsection{Nonlinear Least-Squares Optimization} \label{sec:ls}
The geometric estimates $\mbf{x}_{\text{geom}}^{(i)}, i = 1,\ldots,M$ are used to initialize a nonlinear least-squares algorithm \cite{BarShalom2002EWA} by solving
\begin{align}
\mbfhat{x}_0 = \frac{1}{2} \arg \min_{\mbf{x}} \norm{\mbf{e}(\mbf{x})}^2, \quad \text{where } \quad \mbf{e}(\mbf{x}) = \mbf{g}(\mbf{x}) - \mbfbar{y}. \nonumber
\end{align}
Here, instead of a single set of inter-tag range measurements, an average of $\gamma \geq 100$ range measurements, $\mbfbar{y}$,  are used, which are collected when the robots are static. The averaging enhances the signal-to-noise ratio and improves estimation accuracy. For $i = 1,\ldots, M$, $\mbf{x} \in \SE{2}^{N-1}$ is iteratively updated using the $\oplus$ operator as,
\begin{align}
    \mbfhat{x}^{(i)}_t = \mbfhat{x}^{(i)}_{t-1} \oplus \left(\lambda \; \delta \mbf{x}^{(i)}_{t-1}\right),
\end{align}
where $\lambda$ is the step size, $t$ is the iteration number, and $\mbfhat{x}^{(i)}_0~=~\mbf{x}^{(i)}_{\text{geom}}$.
The optimal step $\delta \mbf{x}^{(i)}_{t-1}$ is given by
\begin{align}
    \label{eq:delta_x}
    \delta\mbf{x}^{(i)}_{t-1} = -\left(\mbf{H}(\mbf{x})^\trans \mbf{H}(\mbf{x})\right)^{-1}\mbf{H}(\mbf{x})^\trans \mbf{e}(\mbf{x})\Bigr|_{\mbfhat{x}^{(i)}_{t-1}},
\end{align}
where $\mbf{H}(\mbf{x})$ is the measurement model Jacobian. The iterations are repeated until $||\delta \mbf{x}^{(i)}_{t-1}||$ is small. In the measurement Jacobian, by taking the measurements between all the tags into account, the least-squares method produces a far more accurate estimate of the ambiguous relative poses than the geometric method. It even reduces the number of ambiguities since it is fed more inter-robot measurement information compared to the geometric method. However, the geometric method confines the initial guesses to a small and informed state space, essential for efficient convergence of the least-squares estimator. 

The covariance of the least-squares estimate represents the uncertainties associated with estimating the state using the range measurements. Assuming that the average range measurements, $\mbfbar{y}$, are unbiased, using this matrix as the covariance of state estimates is a good starting point for any filter initialization. This covariance is given by \cite{Hastie01SL}
\begin{align}
    \mbf{P}_{\tau}^{(i)} = \Sigma^{(i)}(\mbf{H}(\mbfhat{x}^{(i)}_\tau)^\trans\mbf{H}(\mbfhat{x}^{(i)}_\tau))^{-1}, \;\Sigma^{(i)} = \frac{1}{L}\mbf{e}(\mbfhat{x}^{(i)}_\tau)^{\trans}\mbf{e}(\mbfhat{x}^{(i)}_\tau), \nonumber
\end{align}
where $L = |\mc{E}| -(N - 2)$, $\mbf{e}(\mbfhat{x}^{(i)}_\tau) = \mbfbar{y} - \mbf{g}(\mbfhat{x}^{(i)}_\tau)$, $i = 1,\ldots, M$, and $\tau$ is the last iteration number. Finally, the relative poses between Robots 1 and $p$ in the state $\mbf{x}^{(i)}_\tau$ are transformed from $SE(2)$ to $SE(3)$ by augmenting them with zero quantities such that the relative roll, pitch and height are zero, which is a reasonable assumption for the start-up phase, where the robots are at ground level. These estimates and their covariances, denoted as $\{\mbfhat{x}_\tau^{(i)}, \mbfhat{P}_\tau^{(i)}\}_{i=1}^M$, are referred to as the \emph{geometrically-initialized least-squares} (GI-LS) estimates.

This approach is validated in simulation and experiment, as shown in Fig.~\ref{fig_3}, using the problem setup in Fig.~\ref{fig_1}. Here, the lighter shaded drones depict the estimates with their covariances. Using $4\,\si{s}$ of noisy range measurements at $50\,\si{Hz}$ in simulation and $5\,\si{s}$ at $90\,\si{Hz}$ in experiment, both with a covariance $\mbf{R} = 0.1^2\mbf{1}\,\si{m}^2$, the proposed method identifies all four ambiguities in $\SE2$ for each robot. The estimates are accurate despite noise and disturbances. Furthermore, the estimates with lower covariances are more likely to be the ``true'' mode, given that the covariance indicates confidence. Note that, since the least squares method has more measurement information, in Fig.~\ref{fig_3}, it is able to reduce the number of ambiguities from $16$ geometric estimates to $8$ final estimates for the three-robot scenario.

\section{Simulations}\label{sec:simulation} 

\begin{figure}[t]
    \vspace{-0.02cm}
    \centering
    \subfloat{{\includegraphics[width=0.98\columnwidth, trim={0cm 0cm 0.2cm 0cm}, clip] {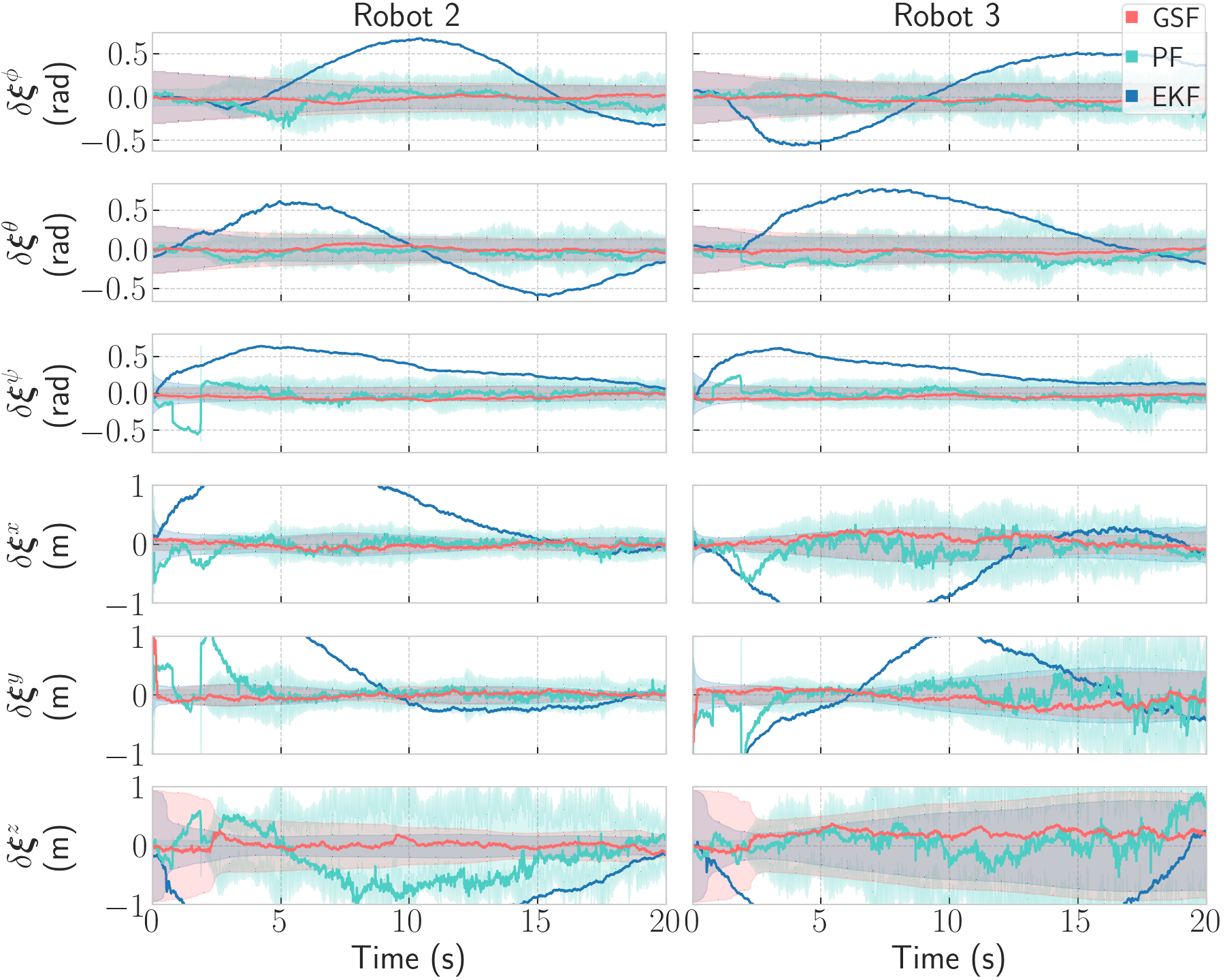}}}%
    \caption{The performance of the EKF, GSF and PF on simulated data for two-tag Robots $2$ and $3$, with Robot~$1$ as reference robot. The GSF and PF are initialized with $8$ GI-LS estimates and $1500$ particles, respectively. The EKF is initialized in a wrong mode among the 8 GI-LS estimates. The shaded regions represent the $\pm 3\sigma$ bounds.}%
    \label{fig_sim}
\end{figure}

\begin{figure}[!t]
    \centering
    \includegraphics[width=0.93\columnwidth, trim={10.4cm 0.8cm 10.9cm 1.6cm}, clip]{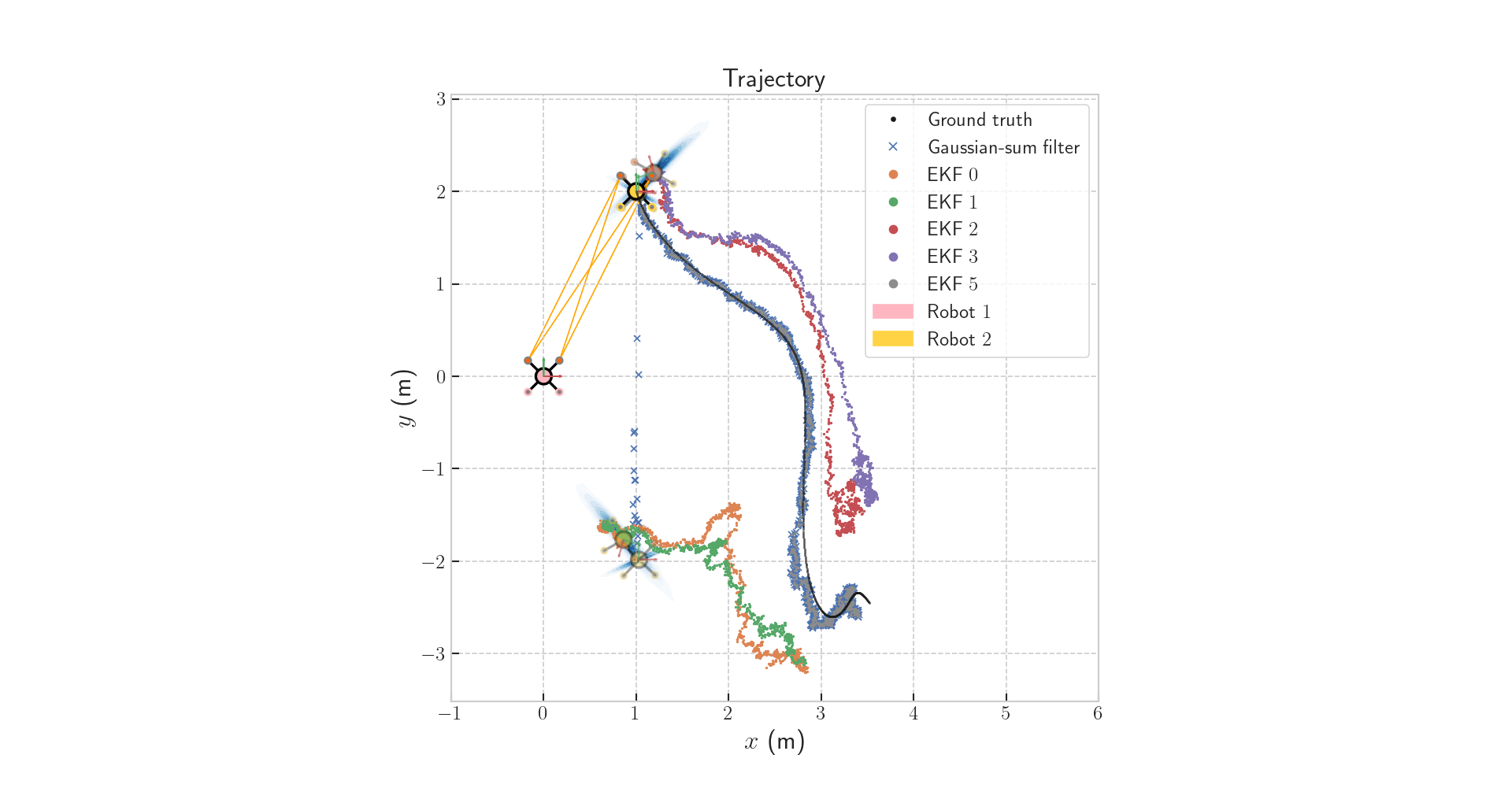}
    \caption{GSF trajectory estimation plot for a single run in simulation, shown in 2D. Only some modes of the GSF and only the relative position between Robot$~1$ and Robot$~2$ are shown for clarity. The ground truth starts at the location the quadcopters are plotted, and Robot~$1$ is the reference robot.}
    \label{fig_6}
    \vspace{-0.5cm}
\end{figure}

\begin{figure}[t]
    \centering
    \includegraphics[width = 3.2in, trim={0cm 0cm 0cm 0.1cm}, clip]{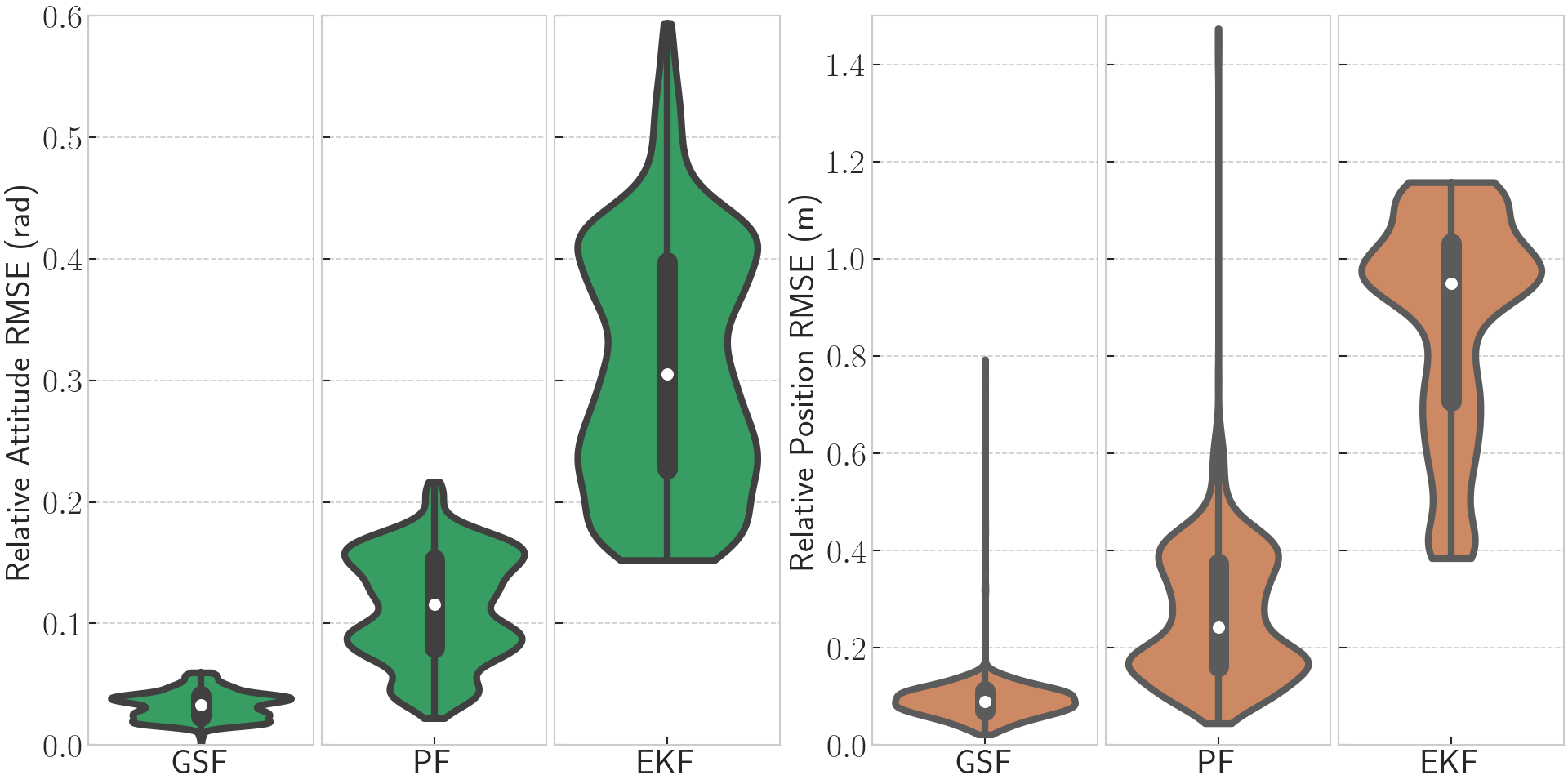}
    \caption{Violin and box plots showing the distribution of the 100-trial attitude and position RMSEs for simulation in $SE(3)$. The envelope shows the relative frequency of RMSE values. The white dot is the median, and the lower and upper bound of the black bar represent the first and third quartile of the data, respectively.}
    \label{fig_5}
\end{figure}

The GSF with its proposed initialization features is compared with the PF and EKF in simulation. The setup is shown in Fig.~\ref{fig_1}, where the three robots have two tags each, and Robot~$1$ is the reference robot. The two tags are located at 
\begin{align}
    \mbf{r}^{\tau_ip}_p = [0.17 \;\;0.17 \;\;0]^\trans \text{ and } \mbf{r}^{\tau_jp}_p = [0.17 \;\;-0.17 \;\;0]^\trans, \nonumber
\end{align}
where $i$ and $j$ are the tag IDs, $p$ is the robot ID, and the units are in meters. The robot velocities are inputs to the process model, and inter-tag range data at $50\,\si{Hz}$, with a covariance of $\mbf{R} = 0.1^2\mbf{1}\,\si{m}^2$ are the measurements.

In Fig.~\ref{fig_sim}, the pose-error plots for a single run of the GSF, PF, and EKF in simulation are shown. The GSF is initialized with $8$ equally-weighted GI-LS estimates, $\{\mbfhat{x}^{(i)}_\tau, \mbfhat{P}^{(i)}_\tau\}_{i=1}^{8}$, the PF with $1500$ particles around the ambiguities, and the EKF is initialized in a wrong mode among the $8$ GI-LS estimates. From the error plots alone, for this single run, despite the GSF having far fewer Gaussian components than the PF's particles, it is visibly more stable and accurate. The EKF diverges since it is initialized in a wrong mode. An EKF initialized in the correct mode is not shown, as it is not practical to know the correct mode in real-world scenarios. Additionally, Fig.~\ref{fig_6} shows the GSF trajectory estimation plot in 2D for the same run. For clarity of reading the plot, only the relative position estimates between Robot~$1$ and Robot~$2$ and only some modes of the EKFs running inside the GSF are shown. The plot clearly shows that the GSF almost instantaneously converges to the ``true'' EKF as its highest-weighted mode, which is EKF~$5$.

\begin{figure}[t]
    \centering
    \includegraphics[width=\columnwidth, trim={0cm 0cm 0.1cm 0cm}, clip]{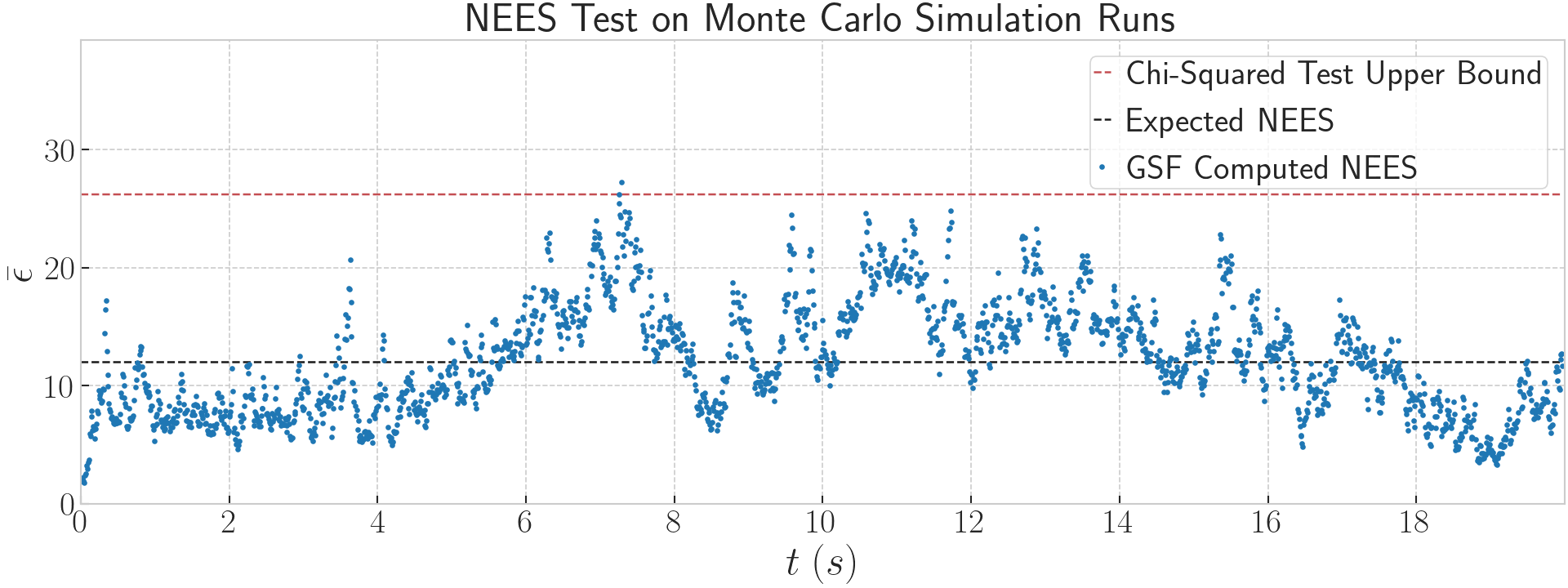}
    \caption{100-trial NEES plot for the proposed GSF estimator in simulation.}
    \label{fig_4}
    \vspace{-0.3cm}
\end{figure}

The proposed GSF's performance is assessed over $100$ Monte-Carlo trials with varied initial conditions and noise realizations on random trajectories. Its \emph{root-mean-squared error} (RMSE) is compared to $100$ EKF and PF trials. The GSF is initialized with $8$ GI-LS estimates, and the PF with $1500$ particles, and the EKF is randomly initialized in one of the $8$ modes. In Fig.~\ref{fig_5}, the GSF has a median attitude RMSE of $0.034\,\si{rad}$, which is $70.6\%$ lower than the PF's $0.116\,\si{rad}$, and EKF's $0.305\,\si{rad}$. Similarly, the median position RMSE is $0.090\,\si{m}$ for GSF, $0.242\,\si{m}$ for PF and $0.949\,\si{m}$ for EKF. Due to the proposed initialization method, the Gaussian components are highly informative while being far fewer than the particles in PF. This allows the GSF to converge to the true mode faster than the PF, making it more accurate, and computationally more efficient. The normalized estimation error squared (NEES) test \cite[Ch. 5.4]{BarShalom2002EWA} in Fig.~\ref{fig_4} confirms GSF's consistency within a 99\% confidence interval. 
\section{Experimental Results}\label{sec:experimental}
\begin{figure}[h]
    \centering
    \includegraphics[width = \columnwidth, trim={0cm 0.cm 0cm 0cm}, clip]{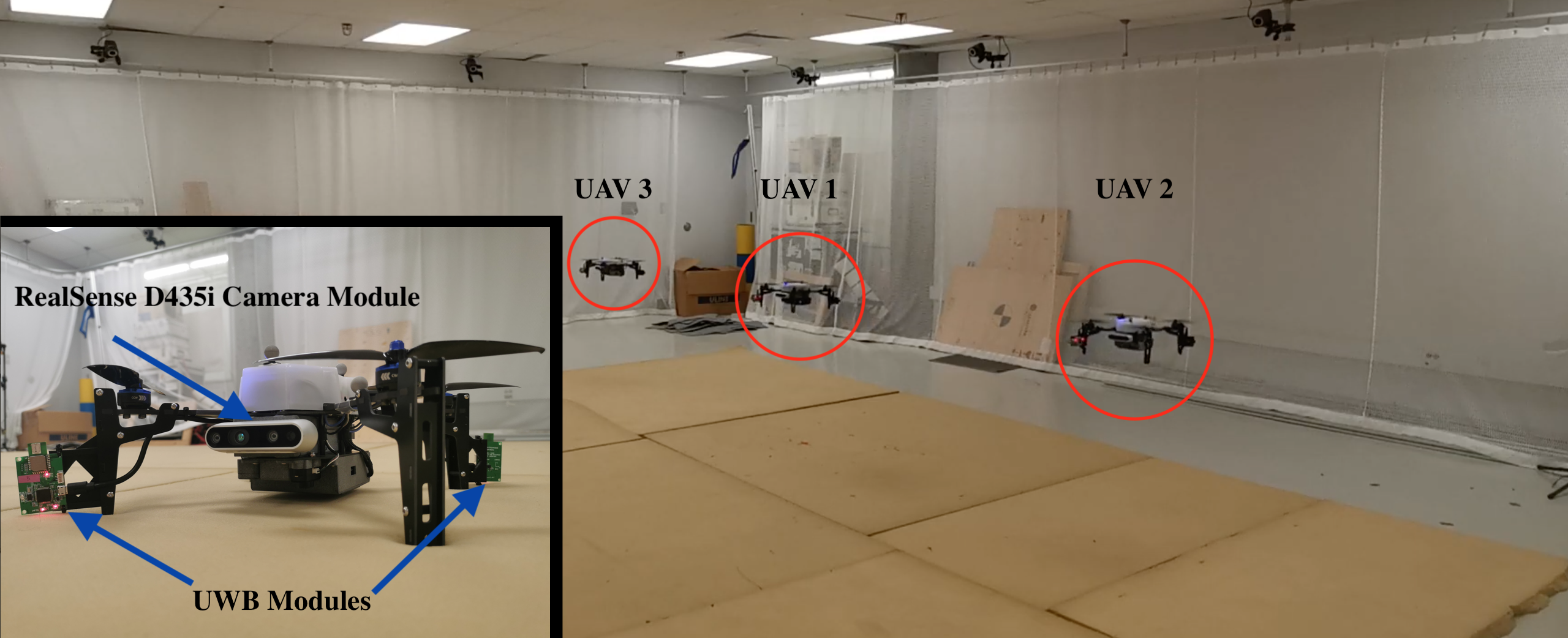}
    \caption{Experimental setup showing the three robots. Two UWB modules or tags and an Intel RealSense D435i camera are mounted on each robot.}
    \label{exp}
\end{figure}
The filters are tested on three Uvify IFO-S quadcopters to validate their performances in experiment. The setup of three robots is depicted in Fig.~\ref{fig_1}, with each robot having two tags, and Robot~$1$ is the reference robot. The two tags in all the robots are located at
\begin{align}
    \mbf{r}^{\tau_ip}_p = \bbm 0.16\\ -0.17\\ -0.05\ebm,
    \mbf{r}^{\tau_jp}_p = \bbm -0.17\\ 0.16\\ -0.05 \ebm, \nonumber
\end{align}
where $i$ and $j$ are the tag IDs, $p$ is the robot ID, and the units are in meters. Each robot has an onboard IMU and an Intel RealSense D435i stereo camera set. These sensors provide the translational velocity estimates through VIO using the ROS package Vins-Fusion \cite{qin2017vins} at $30\,\si{Hz}$, and the angular velocity readings are taken from the gyroscope at $200\,\si{Hz}$. The velocity estimates from VIO only serve as interoceptive measurements to validate the proposed estimation approach. Any other interoceptive measurements can be used in place of VIO as well. Pose data from the Vicon motion-capture system serve as ground truth. The robots follow a random 3D trajectory in a $6\times 6 \times3\,\si{m}^3$ space as shown in Fig.~\ref{exp}. 

\begin{figure}[h]
    \centering
    \subfloat{{\includegraphics[width=\columnwidth, trim={0cm 0cm 0cm 0cm}, clip]{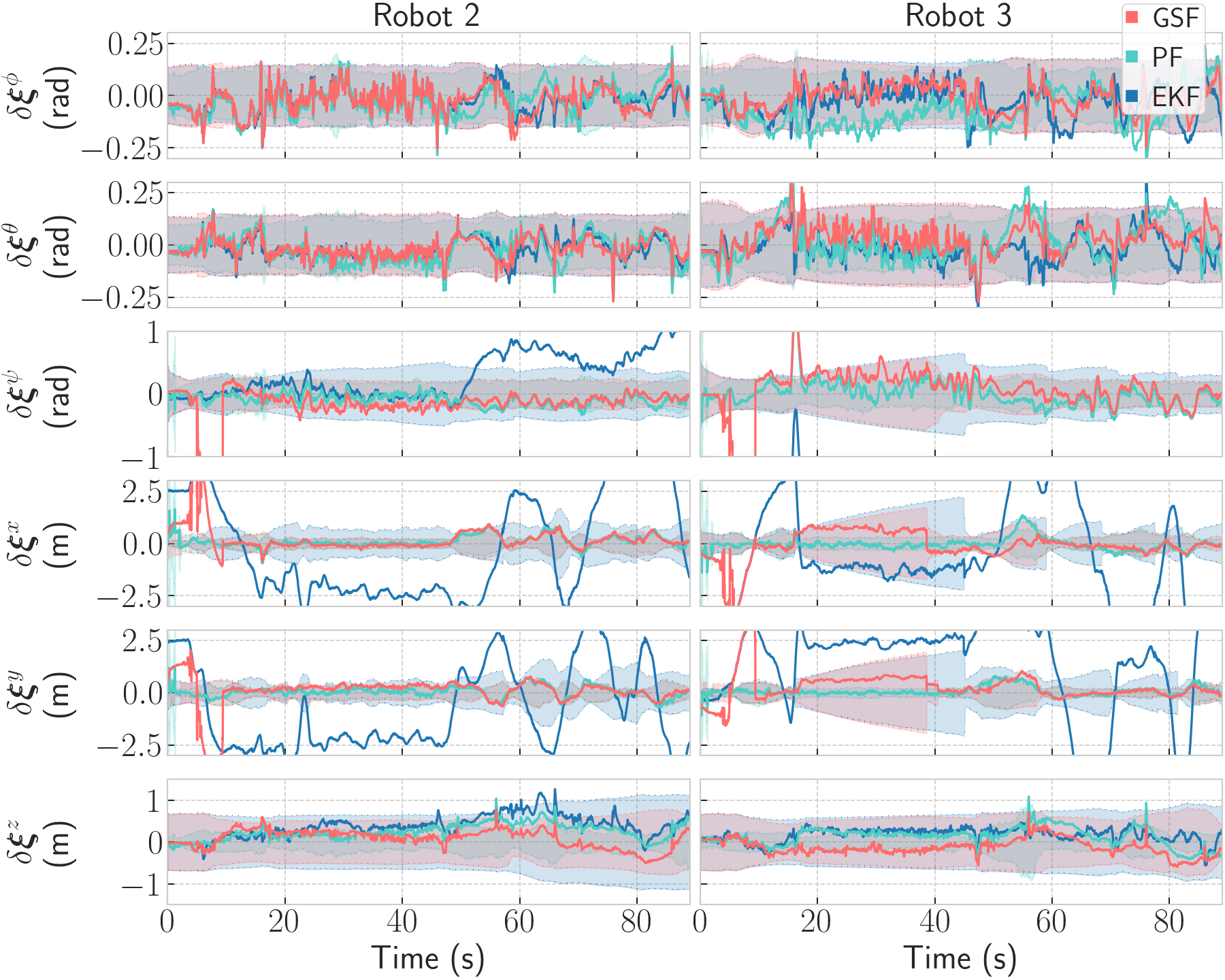} }}%
    \caption{The performance of the EKF, GSF and PF on experimental data for two-tag Robots $2$ and $3$, with Robot~$1$ as reference robot. The GSF and PF are initialized with $8$ GI-LS estimates and $1500$ particles, respectively. The EKF is initialized in a wrong mode among the $8$ GI-LS estimates. The shaded regions represent the $\pm 3\sigma$ bounds.}%
    \label{fig_exp}
    \vspace{-0.1cm}
\end{figure}

The UWB range measurements are provided to all the estimators at $90\,\si{Hz}$, which are corrected for uncertainties and biases using the works of \cite{Shalaby2023Calibration}. The GSF is initialized with $8$ Gaussian GI-LS estimates, the PF with $1500$ particles, and the EKF is initialized in a wrong mode among the $8$ GI-LS estimates. Fig.~\ref{fig_exp} displays the pose-error plots of the filters in experiment. The GSF and PF perform similarly, but the EKF diverges as expected. Initially, the GSF has large error spikes, but it soon stabilizes once it isolates the ``true'' mode. In Python $3.8$, the GSF estimates the states at an average rate of $40\,\si{Hz}$, and the PF does the same at $3.5\,\si{Hz}$, making the GSF many folds faster and strongly eligible for online implementation.




\section{Conclusion}\label{sec:conclusion}

Multi-robot systems with non-stationary range sensors suffer from ambiguous poses due to observability issues. This paper provides a complete and efficient 3D relative pose estimation solution for these systems where UWB ranging tags are the only exteroceptive sensors. In simulations and experiments, the proposed estimator in the form of a Gaussian-sum filter is shown to be accurate and comparatively faster than the particle filter. The results establish that a well-modelled GSF should be the default tool for range-based 3D relative pose estimation in multi-robot systems. This, however, is a centralized estimator and having large systems with only one reference robot leads to many Gaussian components, increasing computational complexities. Therefore, for larger systems, a decentralized approach where multiple robots share their pose estimation with their neighbours, the number of Gaussian components in the GSF can be optimized, reducing the computational burden while retaining accuracy.

{\AtNextBibliography{\small}
\printbibliography}
\end{document}